\documentclass{article}

\usepackage{PRIMEarxiv}

\usepackage[utf8]{inputenc} 
\usepackage[T1]{fontenc}    
\usepackage{hyperref}       
\usepackage{url}            
\usepackage{booktabs}       
\usepackage{amsfonts}       
\usepackage{nicefrac}       
\usepackage{microtype}      
\usepackage{lipsum}
\usepackage{fancyhdr}       
\usepackage{graphicx}       
\usepackage{tikz}
\usepackage{pgfplots}
\graphicspath{{media/}}     

\title{Evaluating Large Language Models on Multiword Expressions in Multilingual and Code-Switched Contexts
\thanks{\textit{\underline{Citation}}: 
\textbf{Authors. Title. Pages.... DOI:000000/11111.}} 
}

\author{
  Frances Laureano De Leon \\
  School of Computer Science \\
  University of Birmingham \\
  Birmingham, UK\\
  \texttt{\{fxl846@student.bham.ac.uk} \\
  \And
    Harish Tayyar Madabushi \\
  Department of Computer Science \\
  University of Bath \\
  Bath, UK\\
  \texttt{htm43@bath.ac.uk} \\
   \And
  Mark G. Lee \\
  School of Computer Science \\
  University of Birmingham \\
  Birmingham, UK\\
  \texttt{m.g.lee@bham.ac.uk} \\
}

\begin{document}
\maketitle

\begin{abstract}
Multiword expressions, characterised by non-compositional meanings and syntactic irregularities, are an example of nuanced language. 
These expressions can be used literally or idiomatically, leading to significant changes in meaning. 
While large language models have demonstrated strong performance across many tasks, their ability to handle such linguistic subtleties remains uncertain.
Therefore, this study evaluates how state-of-the-art language models process the ambiguity of potentially idiomatic multiword expressions,  particularly in contexts that are less frequent, where models are less likely to rely on memorisation.
By evaluating models across in Portuguese and Galician, in addition to English, and using a novel code-switched dataset and a novel task, we find that large language models, despite their strengths, struggle with nuanced language. 
In particular, we find that the latest models, including GPT-4, fail to outperform the xlm-roBERTa-base baselines in both detection and semantic tasks, with especially poor performance on the novel tasks we introduce, despite its similarity to existing tasks. Overall, our results demonstrate that multiword expressions, especially those which are ambiguous, continue to be a challenge to models. 
\end{abstract}


\section{Introduction}
\label{intro}
Multiword expressions (MWEs) present significant challenges in Natural Language Processing (NLP) due to their linguistic complexities.
MWEs consist of multiple words that convey a specific meaning, which may be non-compositional and exhibit syntactic irregularities~\cite{Baldwin2010, Constant2017}.
These expressions are widespread across languages and domains, from the arts to the sciences~\cite{Villavicencio2019}.
The number of MWEs in a speaker’s lexicon is estimated to be comparable to that of single words~\cite{Sag2002MultiwordNLP}.
MWEs are often rooted in cultural context and communication rather than a straightforward lexical meaning, making their interpretation in NLP systems more challenging, particularly in idiomatic contexts~\cite{Masini2019Multi-WordMorphology}.
Additionally, MWEs can be ambiguous, functioning either idiomatically or literally, which can cause potential confusion~\cite{kurfal2020}.
For example, the English MWE \textit{break the ice} can refer to its literal meaning or describe making unfamiliar individuals feel more at ease with one another.
Such expressions, known as potentially idiomatic expressions (PIEs)~\cite{Haagsma2020}, require an understanding of both literal and figurative meanings.
Given the idiosyncratic nature of MWEs, it is essential that state-of-the-art NLP models accurately capture their meanings.

Large language models (LLMs) have rapidly become the dominant approach in NLP, demonstrating strong language understanding and generation capabilities across a wide range of tasks in zero-shot and few-shot settings~\cite{Liu2022TestingLanguage}.
However, despite their strong performance, LLMs have limitations, including the generation of hallucinations—fluent but inaccurate or fabricated text—and a reliance on statistical co-occurrence patterns from pre-training data~\cite{razeghi2022, kang2023}.
\emph{This study, therefore, aims to assess whether LLMs can genuinely understand nuanced language, using MWEs as a means of evaluation.}
MWEs are particularly relevant because their meanings can be either idiomatic or literal, requiring an understanding of both social and linguistic aspects of language.
Since MWEs exist in all languages and vary in frequency and usage, their interpretation may be further complicated in multilingual contexts.
One such challenge arises in code-switching (CS), where speakers alternate between languages within a conversation or sentence~\cite{Joshi1982, Dogruoz2021}.
Code-switching is common in multilingual communities and presents additional difficulties for NLP systems, as MWEs may be split across languages.
To evaluate how LLMs handle this complexity, we use data in English, Portuguese, and Galician, as well as CS examples.
By including both high-resource and low-resource languages alongside CS data, we aim to assess whether models rely primarily on statistical patterns in training data or if they can generalise effectively across multilingual and code-switched contexts.

We conduct experiments using both open-source and closed-source LLMs to examine how these models handle MWEs across different languages and contexts. 
Specifically, we investigate:
\begin{itemize}
    \item the ability of LLMs to detect whether MWEs are used as idioms or literally,
    \item the extent to which LLMs effectively capture the non-compositional meanings of MWEs,
    \item the ability of LLMs to acquire information from the prompt, and 
    \item the relationship between LLM size and performance on text containing MWEs with non-compositional meaning.
\end{itemize}

Our experiments reveal that, despite their proficiency in many other tasks, LLMs continue to struggle with processing MWEs. 
This gap highlights the limitations of these models compared to fine-tuned pre-trained language models (PLMs) such as xlm-roBERTa, which are much smaller models and contain many fewer parameters. 
While some results demonstrate handling of MWEs with non-compositional meanings in English comparable to fine-tuned PLMs, the same cannot be said for other languages. 
Our experiments highlight the limitations of current models and emphasise the need for careful consideration when applying them in multilingual contexts, particularly with nuanced language.
Section~\ref{related_work} reviews the relevant literature on MWEs. 
In section~\ref{datasets}, we describe the datasets used in this study and outline the creation of the synthetic CS dataset. 
Section~\ref{experiments} provides a detailed account of our experimental setup, while section~\ref{results} presents and discusses our findings. 
Finally, we conclude with a summary of our results and recommendations for future research in section~\ref{conclusion}.

\subsection{Contributions}
We carry out experiments to assess the ability of generative models to detect and interpret MWEs in both idiomatic and compositional contexts.
MWEs are used for evaluation due to their lower frequency, reducing the likelihood of LLMs relying on memorisation.
To further investigate whether LLMs use memorisation in MWE interpretation, we create a novel CS test dataset with examples in three language pairs—Galician-Spanish, English-Spanish, and Portuguese-Spanish—and make it publicly available.
To address the research questions above, we evaluate models not only on the standard task of detecting multiword expressions (Section \ref{idiom-detection}) but also on novel tasks assessing the semantic interpretation of MWEs (Section \ref{semantics}). This includes a combination of synthetic MWEs (Section \ref{synthetic_MWEs}) and a new code-switched dataset (Section \ref{sec:code-switched}).
By introducing these datasets, tasks, and settings, we aim to reduce the likelihood of model memorisation and assess their ability to interpret MWEs, particularly in non-compositional contexts.
Our results demonstrate that LLMs are not fully capable of detecting and representing the semantics of MWEs. 
Furthermore, they indicate that prompt-based learning, by providing information in the prompt, is not sufficient for the model to handle unseen vocabulary, in challenging linguistic tasks.

\section{Related work}
\label{related_work}
Studies have shown that popular transformer models generally struggle to effectively handle figurative language. 
Research on encoder-only models shows that PLMs do not adequately capture and represent the meanings of idiomatic expressions~\cite{Garcia2021ProbingModels}. 
Additionally, generative models like GPT-2 have been found to perform poorly in handling figurative language without the use of mitigation strategies~\cite{Jhamtani2021InvestigatingConstructs}.
\cite{Miletic2024SemanticsSurvey} reveal that transformer models capture MWE semantics inconsistently, and find that they are reliant on surface patterns.
However, research has shown that encoder-only PLMs can effectively represent MWEs when fine-tuned with appropriate data~\cite{Madabushi2022}.
Studies exploring how generative models handle figurative language are few, especially studies focusing on idioms.
\cite{Liu2022TestingLanguage} test BERT, and RoBERTa, as well as generative models, namely, GPT-2, GPT-3 and GPT-NEO on their ability to reason about figurative language, with a focus on metaphors.
They find that all models need to be fine-tuned to do well on interpreting figurative language, and that the capabilities of these models remain far from reaching human performance.
\cite{DeLucaFornaciari2024AModels} conduct experiments on an English language dataset they create to assess model abilities in detecting idiomatic expressions in a zero-shot setting. 
They conduct their experiments on the 7B versions of Llama-2, Vicuna, and Mistral models.
\cite{Phelps2024SignDetection} conduct detection experiments in English, Portuguese and Galician on different state-of-the-art generative models and find that although they give competitive results, they fail to reach the performance of fine-tuned PLMs.
In this work, we will also explore model abilities in detecting MWEs in English, Portuguese, and Galician.
However, our work also examines how generative models represent the \textit{meaning} of sentences where MWEs are used idiomatically in these languages.
Additionally, we investigate the impact of CS text on model performance and assess the models' ability to manage unseen MWEs using synthetic examples.

\section{Datasets}
\label{datasets}
To assess the capabilities of LLMs in understanding nuanced language, we use the SemEval 2022 Task 2 dataset~\cite{Madabushi2022} for all experiments.
This dataset contains MWEs in both literal and idiomatic contexts in English, Portuguese, and Galician, and includes adversarial examples to test model consistency in interpreting idiomatic meaning.
The SemEval task comprises two subtasks: \textit{subtask a}, an MWE interpretation task, and \textit{subtask b}, an MWE paraphrase similarity task adapted from a semantic text similarity (STS) task for use with generative models.
Both tasks are designed to assess whether models can capture the meaning of sentences containing MWEs, whether used compositionally or not.
We adapt these subtasks and create new tasks based on the same data to test whether familiarity with task format—likely present in instruction tuning—affects performance.
This also motivates our use of variations on a single dataset rather than multiple datasets.
A further advantage of this dataset is its multilingual nature and the fact that the test labels have not been released, reducing the chance that models have seen the answers during instruction tuning or pre-training.

\subsection{Code-Switched Dataset}
\label{sec:code-switched}
To further evaluate the models' ability to interpret text containing MWEs, we create a novel code-switched (CS) dataset based on the previously described data.
CS commonly occurs in multilingual communities worldwide and contributes to the development of mixed language varieties such as Spanglish (a Spanish–English blend).
This synthetic CS dataset was created by combining Spanish with the original English, Portuguese, and Galician monolingual sentences.

We generate the CS examples using OpenAI’s GPT-4-turbo-0125 model with a temperature setting of 0.65 and a seed value of 42.
A temperature of 0.65 offers a balance between determinism and diversity, allowing the model to produce varied yet coherent outputs while ensuring consistent incorporation of Spanish within sentence structures.

Two base prompts were used in the generation process: one for all examples containing MWEs (from either subtask a or subtask b), and another for the adversarial examples in subtask b, which do not include MWEs but are paraphrases of their literal meanings.
Generation was carried out separately for each language pair, mixing Spanish individually with English, Portuguese, and Galician, to minimise errors in the outputs.
The prompts used are shown in Table~\ref{tab:CS-prompts}.

Understanding how MWEs function in a CS context offers insight into modelling real-world language use.
As generative models are increasingly used by multilingual speakers, this study evaluates how well state-of-the-art models such as GPT handle mixed-language input.
This approach also highlights challenges that may be overlooked by monolingual evaluation methods.
Thus we Use CS text to help ensure that sentence meanings remain consistent with the original dataset while reducing the likelihood that models rely on surface-level co-occurrence patterns.

\section{Experimental Setup}
\label{experiments}
In this section, we introduce our experiments and the models used as part of this work.
We run three different experiments: an~\textbf{MWE interpretation task}, a~\textbf{synthetic MWE interpretation task}, and~\textbf{MWE paraphrase similarity task}. 
These experiments aim to determine whether the models can recognise the presence and meaning of text that contains nuanced language.
Most of the experiments are conducted in a zero-shot and few-shot setting. 

\subsection{Models}
\label{models}
We use a total of four models for experiments, both open-source and closed-source models:
(1) gpt-3.5-turbo-0125, (2) gpt-4-0125-preview, (3) meta-llama-3-70b-instruct, and (4) meta-llama-3-8b-instruct.
We utilise the OpenAI API\footnote{\url{https://openai.com/}} for our experiments on the GPT models and use the Replicate API\footnote{\url{https://replicate.com/}} for our experiments on the Llama models.
For all models, we use the following parameters: temperature 0, and seed 42.
For the Llama models we specify a max token count of 7 and for the GPT models we use a count of 5.

\subsection{Prompts}
\label{prompts}
For all experiments in this section, we use three different prompts to generate outputs from the LLMs.
The prompts are based on strategies outlined on the OpenAI website~\footnote{\url{https://platform.openai.com/docs/guides/prompt-engineering/six-strategies-for-getting-better-results}}. 
Specifically, we employ three approaches: the 'persona' prompt, where the model adopts the role of a linguist; the 'think' prompt, where the model is instructed to reflect on the meaning of a sentence before assigning a label; the 'ChatGPT' prompt, generated using ChatGPT-turbo-3.5. 
Each prompt is adapted according to the specific task.
We opt to use three prompts because model results may vary given different inputs~\cite{Liu2024AreSayings}, and this allows to explore possible variations in the model outputs given variations in the input.
The prompts are available in Table~\ref{tab:prompts-experiments}.

\subsection{MWE Interpretation Task}
\label{idiom-detection}
We conduct MWE sense disambiguation experiments, in which we ask an LLM to determine whether a given MWE in a sentence is used idiomatically or literally.
We carry out these experiments in both zero-shot and few-shot settings using data from subtask a of the SemEval competition, which includes a total of 2,342 examples: 916 in English, 713 in Portuguese, and 713 in Galician.
For the few-shot experiments, the model is provided with five examples: three in English and two in Portuguese. 
At least one instance of an MWE being used literally and idiomatically is included for each of the above languages. 
The original test dataset for SemEval subtask a, along with the synthetic CS dataset, is used for this task.
The languages of the few-shot examples is not altered for the CS experiments, that is to say, the examples provided to the LLMs is just in English and Portuguese. 
The model receives input which consists of text containing an MWE, along with the MWE itself.
We prompt the model to generate a label, either 'idiom' or 'literal', in response to the question, 'Is the multiword expression [MWE] an idiom or literal in this sentence?'.
The three base prompts previously mentioned are used to generate answers from each model. 
The results of these experiments are presented in Table~\ref{tab:idiom-detection-zero-shot} and Table~\ref{tab:idiom-detection-few-shot}.
\begin{figure}[ht!]
\centering









\includegraphics[scale=0.5]{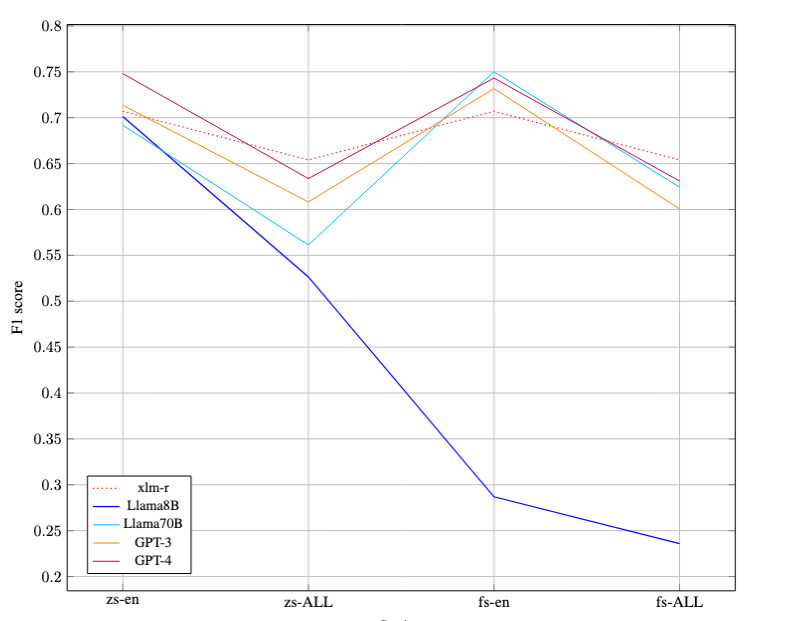}
\caption{F1 scores for each model using the original SemEval task 2 dataset. This graph shows the F1 score for the English language subset of the dataset, as well as the total macro F1 score for all languages combined in the zero-shot (zs) and few-shot settings (fs).}
\label{fig:f1-original-dataset}
\end{figure}

\subsection{Synthetic MWE Interpretation Task}
\label{synthetic_MWEs}
The objective of the synthetic MWE experiments is to assess whether providing additional information in the prompt can improve model performance.
We specifically aim to test whether models can understand nuanced language when given more context.
This approach is based on~\cite{Eisenschlos2023WinoDict:Acquisition}, where synthetic words were introduced and defined as existing concepts to evaluate an LLM's ability to learn new vocabulary through prompting.
It is likely that an LLM has not encountered all the MWEs in our datasets, particularly those in Portuguese and Galician.
Therefore, it is important to evaluate whether a model can acquire knowledge of new words or MWEs through prompt-based learning.

To investigate this, we use the development set of subtask a, which includes examples in English and Portuguese (but not in Galician), and replace the MWEs in each example with synthetic ones.
The model is prompted using the same three base prompts from the MWE interpretation experiments, with an added definition for the synthetic MWE, taken from the original MWE it replaced.
The results of these experiments are shown in Table~\ref{tab:task-a-zero-shot-synthetic}.

\begin{figure}[h]
\centering









\includegraphics[scale=0.5]{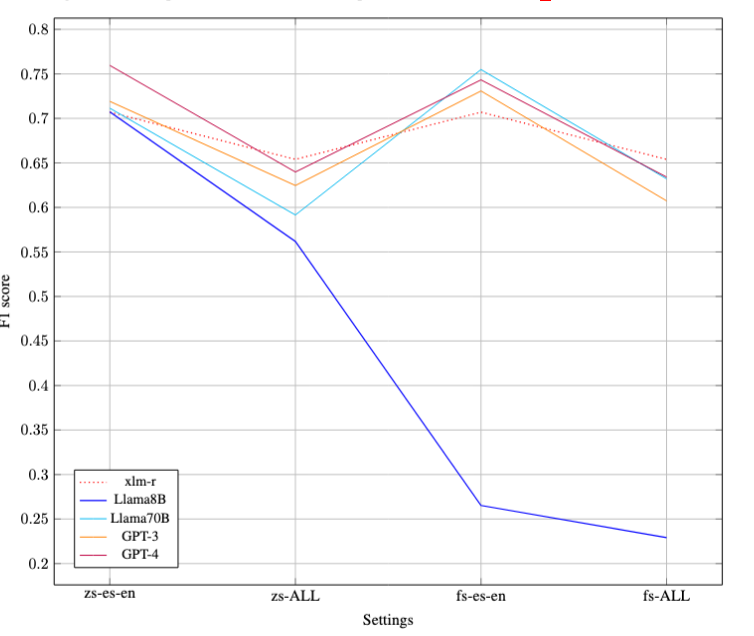}
\caption{F1 scores for each model using the CS dataset. This graph shows the F1 score for the Spanish-English (es-en) language subset of the dataset, as well as the total macro F1 score for all CS examples combined in the zero-shot (zs) and few-shot (fs) settings.}
\label{fig:f1-cs-dataset}
\end{figure}

\begin{figure*}[h]
\center
  \includegraphics[width=0.48\linewidth]{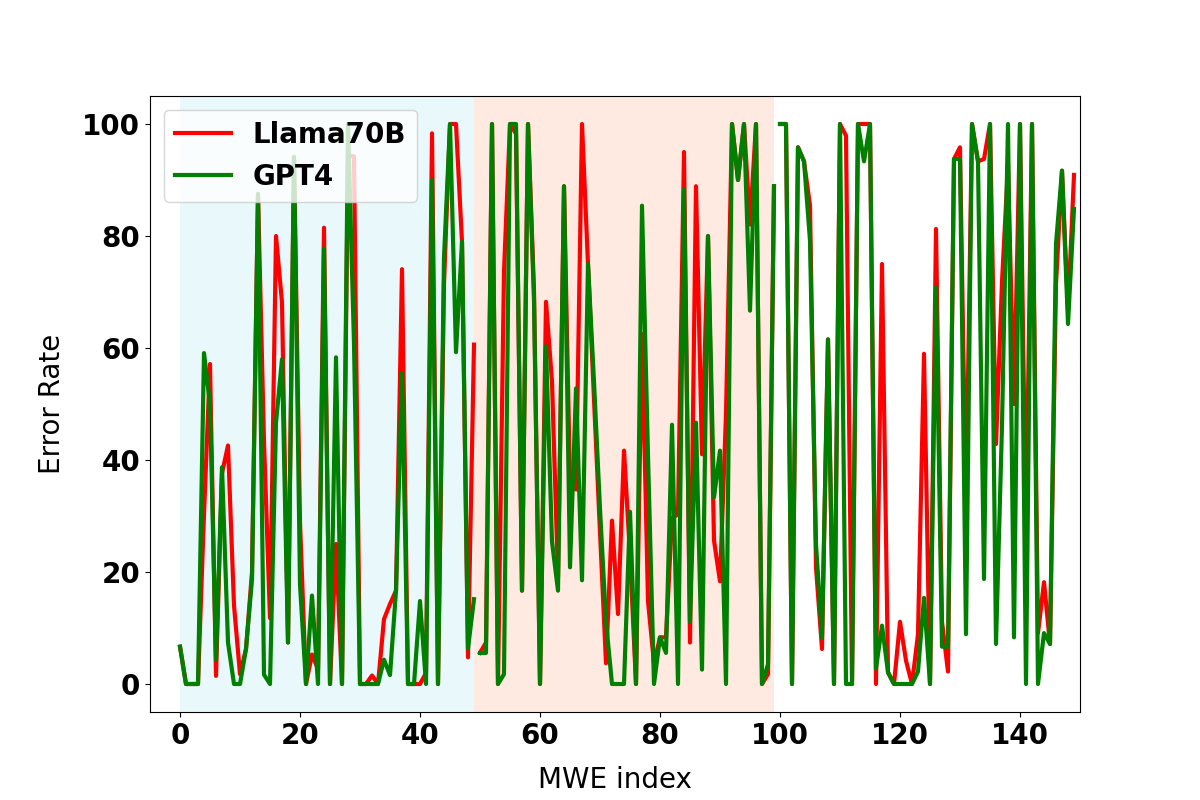} 
  \includegraphics[width=0.48\linewidth]{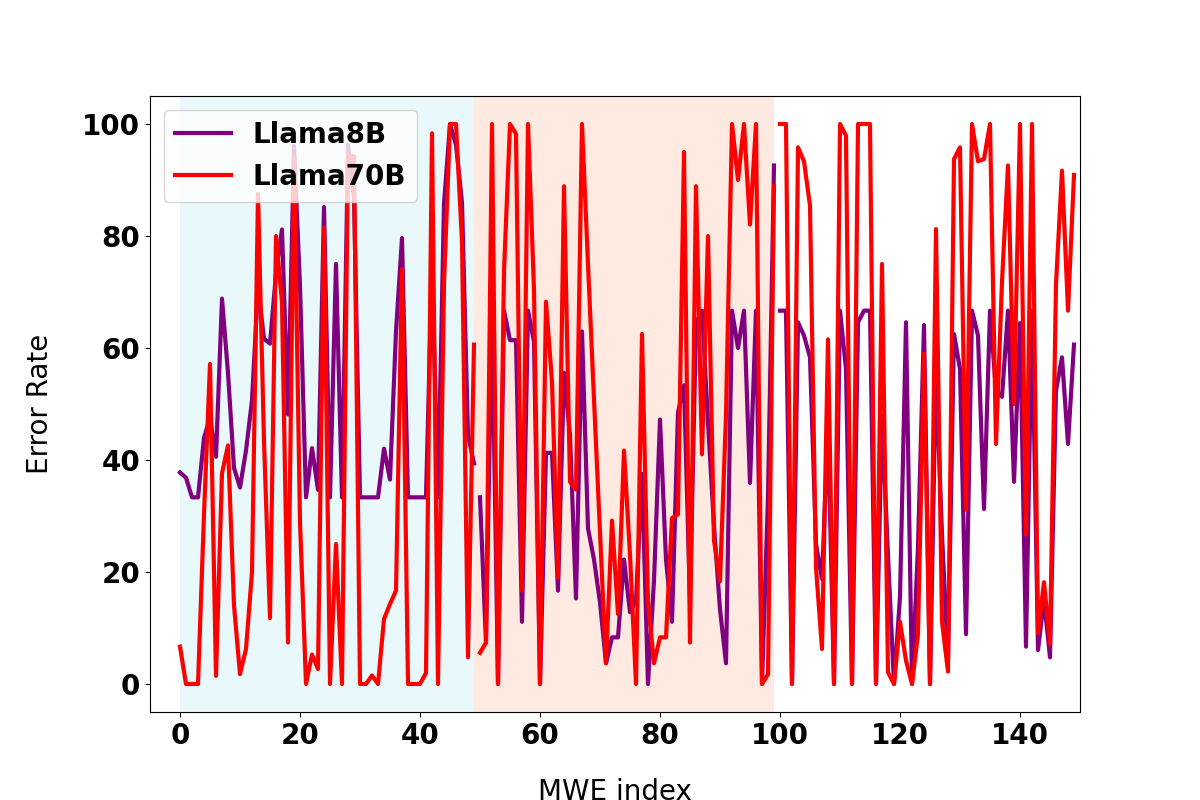}
  \caption {The error rates of all MWEs in the test set. These are in alphabetical order per language. The MWEs in English are in the light blue background, Portuguese are in the light orange background and Galician are in a white background. The figure on the left show the errors for LLama 70B and GPT-4 and the figure on the right shows the errors for the Llama 70B and Llama 8B.}
  \label{fig:errors_mwe}
\end{figure*}
\begin{table*}[h]
\resizebox{\textwidth}{!}{%
\begin{tabular}{lllllll}
\textbf{Model} & \textbf{\begin{tabular}[c]{@{}l@{}}F-1 Score\\ EN\end{tabular}} & \textbf{\begin{tabular}[c]{@{}l@{}}F-1 Score\\ PT\end{tabular}} & \textbf{\begin{tabular}[c]{@{}l@{}}F-1 Score\\ GL\end{tabular}} & \textbf{\begin{tabular}[c]{@{}l@{}}F-1 Score\\ ALL\end{tabular}} & \textbf{Code-switched}  \\ \hline
GPT-4 & $\textbf{0.7480} \pm 0.23$ & $0.5701\pm 0.005$ & $\textbf{0.5395}\pm 0.012$ & $0.6336 \pm 0.013$ & No  \\
GPT-3 & $0.7135 \pm 0.030$ & $0.5359 \pm 0.015$ & $0.5243 \pm 0.101$ & $0.6082 \pm 0.053$ &   \\
Llama 3-70B & $0.6918 \pm 0.025$ & $0.5097 \pm 0.32$ & $0.4420 \pm 0.044$ & $0.5613 \pm 0.034$ &  \\
Llama 3-8B & $0.7013 \pm 0.022$ & $0.4628 \pm 0.047$ & $0.3636 \pm 0.061$ & $0.5265\pm 0.032$ &  \\ \hline
xlm-roBERTta & 0.7070 & \textbf{0.6803} & 0.5065 & \textbf{0.6540} &  &  \\ \hline
GPT-4 & $\textbf{0.7595} \pm 0.025$ & $0.5759 \pm 0.002$ & $0.5411 \pm 0.007$ & $0.6398 \pm 0.012$ & Yes  \\
GPT-3 & $0.7191 \pm 0.021$ & $0.5417 \pm 0.012$ & $\textbf{0.5575} \pm 0.063$ & $0.6246 \pm 0.030$ &   \\
Llama 3-70B & $0.7114\pm 0.020$ & $0.5496 \pm 0.033$ & $0.4726 \pm 0.047$ & $0.5915 \pm 0.034$ &   \\
Llama 3-8B & $0.7072\pm 0.024$ & $0.5276 \pm 0.051$ & $0.4076 \pm 0.040$ & $0.5619 \pm 0.034$ &   \\ \hline
\end{tabular}%
}
\caption{These are the scores for the detection experiments in the \textbf{zero-shot }setting. Reported F-1 scores are for each of the individual languages tested, as well as the combined Macro-F-1 score (ALL).}
\label{tab:idiom-detection-zero-shot}
\end{table*}

\begin{table*}[h]
\resizebox{\textwidth}{!}{
\begin{tabular}{lllllll}
\textbf{Model} &
  \textbf{\begin{tabular}[c]{@{}l@{}}F-1 Score\\ EN\end{tabular}} &
  \textbf{\begin{tabular}[c]{@{}l@{}}F-1 Score\\ PT\end{tabular}} &
  \textbf{\begin{tabular}[c]{@{}l@{}}F-1 Score\\ GL\end{tabular}} &
  \textbf{\begin{tabular}[c]{@{}l@{}}F-1 Score\\ ALL\end{tabular}} &
  \textbf{Code-switched}  \\ \hline
GPT-4        & $0.7432 \pm 0.006$         & $0.5705\pm 0.008$  & $\textbf{0.5356}\pm 0.008$ & $0.6310 \pm 0.003$ & No \\
GPT-3        & $0.7310\pm0.015$            & $0.5455 \pm 0.003$ & $0.4755 \pm 0.092$         & $0.6006 \pm 0.034$ &                  \\
Llama 3-70B  & $\textbf{0.7501} \pm 0.017$         & $0.5638 \pm 0.007$ & $0.5161 \pm 0.016$         & $0.6243 \pm 0.014$ &              \\
Llama 3-8B   & $0.2869 \pm 0.397$         & $0.1864 \pm 0.305$ & $0.1996 \pm 0.341$         & $0.2359 \pm 0.360$   &          \\   \hline
xlm-roBERTta & 0.7070                     & \textbf{0.6803}    & 0.5065                     & \textbf{0.6540}    &             \\ \hline
GPT-4        & $0.7433\pm 0.014$          & $0.5831 \pm 0.012$ & $\textbf{0.5342} \pm 0.002$         & $0.6341 \pm 0.009$ & Yes              \\
GPT-3        & $0.7308 \pm 0.009$         & $0.5433 \pm 0.009$ & $0.4988 \pm 0.064$         & $0.6074 \pm 0.024$ &              \\
Llama 3-70B  & $\textbf{0.7549}\pm 0.011$ & $0.5666 \pm 0.007$ & $0.53 \pm 0.014$           & $0.6321 \pm 0.008$ &             \\
Llama 3-8B   &  $0.2653 \pm 0.391$         & $0.1737 \pm 0.261$ & $0.2107 \pm 0.327$         & $0.2291 \pm 0.346$   &          \\   \hline
\end{tabular}%
}
\caption{These are the scores for the detection experiments in the \textbf{few-shot} setting. Reported F-1 scores are for each of the individual languages tested, as well as the combined Macro-F-1 score (ALL).}
\label{tab:idiom-detection-few-shot}
\end{table*}

\subsection{MWE Paraphrase Similarity Task}
\label{semantics}
As part of this work, we aim to evaluate the ability of LLMs to understand the meaning of text containing MWEs.
We provide each model with sentence pairs: some include an MWE, while others contain correct or incorrect paraphrases of MWEs.
An example of the input provided to the model is shown in Table~\ref{tab:examples_data}.
To run these experiments, we adapt subtask b from the SemEval Task 2 competition, which was originally designed as a STS task.
We construct three distinct prompts, as described in Section~\ref{prompts}, each asking the model whether the two sentences have similar meanings.
Each prompt notes that some sentences may include MWEs used idiomatically.
The model is instructed to output 'true' if the sentences are similar in meaning and 'false' if they differ, turning the task into a binary classification problem.

These experiments are conducted in both zero-shot and few-shot settings.
We use the development split of subtask b from the original SemEval competition, which contains examples in English and Portuguese only.
This dataset includes sentence pairs with MWEs and their paraphrases, as well as examples from a standard STS benchmark.
The gold labels in the dataset are either a Spearman rank correlation score (for STS benchmark examples), a score of 1 (for identical meanings), or a label of NONE (for differing meanings).
After adapting the task and removing examples with similarity scores other than 1 or NONE, we retained 974 examples: 521 in English and 454 in Portuguese.
Since the task requires binary outputs, we exclude all examples with a Spearman score other than 1 and use only examples labelled as 1 (similar) or NONE (not similar).
This required access to the gold labels, and as a result, we were unable to use the original competition evaluator, which assumes unmodified scoring.
In addition to testing on the original development data, we run experiments on our generated CS examples, where each language is mixed with Spanish.
The results for these experiments are shown in Table~\ref{tab:task-b-zero-shot} and Table~\ref{tab:task-b-few-shot}.
We also experiment with the same dataset used for the semantic task.
We substitute the real MWEs in each example with a synthetic MWE and supply the model with the original MWE's definition as the meaning of the synthetic one.
The results of these experiments are on Table~\ref{tab:synthetic-task-b}.

\begin{table*}[ht]
\centering
\begin{tabular}{llllll}
\textbf{Model} &
  \textbf{\begin{tabular}[c]{@{}l@{}}F-1 Score\\ EN\end{tabular}} &
  \textbf{\begin{tabular}[c]{@{}l@{}}F-1 Score\\ PT\end{tabular}} &
  \textbf{\begin{tabular}[c]{@{}l@{}}F-1 Score\\ ALL\end{tabular}} &
  \textbf{Code-switched}  \\ \hline
GPT-4        & $0.7178\pm 0.040$ & $0.6142\pm 0.019$ & $0.6684\pm 0.029$ & No  \\
GPT-3        & $0.6396\pm0.044$  & $0.6574\pm 0.035$ & $0.6490\pm 0.039$ &          \\
Llama 3-70B  & $0.6802\pm 0.071$  & $0.5065\pm 0.050$  & $0.5969\pm 0.061$ &            \\
Llama 3-8B   & $0.6142\pm 0.108$ & $0.4872\pm 0.075$ & $0.5538\pm 0.093$ &         \\ 
xlm-roBERTta & \textbf{0.7590}   & \textbf{0.7658}   & \textbf{0.7712}   &            \\ \hline
GPT-4        & $0.6201\pm 0.023$ & $0.5538\pm 0.003$ & $0.5889\pm 0.012$ & Yes          \\
GPT-3        & $0.5373\pm 0.072$ & $0.5690\pm 0.016$ & $0.5525\pm 0.039$ &               \\
Llama 3-70B  & $0.5012\pm 0.060$ & $0.4189\pm 0.032$ & $0.4618\pm 0.045$ &                \\
Llama 3-8B   & $0.5541\pm 0.019$ & $0.4401\pm 0.018$ & $0.4993\pm 0.018$ &               \\ 
xlm-roBERTta & \textbf{0.6752}   & \textbf{0.7180}   & \textbf{0.7020}   &               \\ \hline
\end{tabular}%

\caption{Results for semantic experiments in the \textbf{zero-shot} setting.}
\label{tab:task-b-zero-shot}
\end{table*}


\begin{table*}[ht]
\centering
\begin{tabular}{llllll}
\textbf{Model} &
  \textbf{\begin{tabular}[c]{@{}l@{}}F-1 Score\\ EN\end{tabular}} &
  \textbf{\begin{tabular}[c]{@{}l@{}}F-1 Score\\ PT\end{tabular}} &
  \textbf{\begin{tabular}[c]{@{}l@{}}F-1 Score\\ ALL\end{tabular}} &
  \textbf{Code-switched} \\ \hline
GPT-4        & $\textbf{0.7628} \pm 0.030$ & $0.6304\pm 0.013$  & $0.7000 \pm 0.021$ & No  \\
GPT-3        & $0.6497\pm0.015$            & $0.6296\pm 0.109$  & $0.6404 \pm 0.060$ &                   \\
Llama 3-70B  & $0.7204\pm 0.006$            & $0.5570\pm 0.006$   & $0.6425 \pm 0.006$ &                 \\
Llama 3-8B   & $0.7045 \pm 0.009$          & $0.5771 \pm 0.010$ & $0.6440\pm 0.008$  &                  \\ 
xlm-roBERTta & 0.7590                      & \textbf{0.7658}    & \textbf{0.7712}    &                \\ \hline
GPT-4        & $0.5455\pm 0.008$           & $0.5209 \pm 0.022$ & $0.5362 \pm 0.014$ & Yes              \\
GPT-3        & $0.5634 \pm 0.051$          & $0.5841 \pm 0.011$ & $0.5732 \pm 0.026$ &                  \\
Llama 3-70B  & $0.5670\pm 0.010$           & $0.4434 \pm 0.004$ & $0.5071\pm 0.007$  &                  \\
Llama 3-8B   & $0.6172 \pm 0.011$          & $0.5073\pm 0.030$  & $0.5651\pm 0.020$  &                  \\ 
xlm-roBERTta & \textbf{0.6752}             & \textbf{0.7180}    & \textbf{0.7020}    &                 \\ \hline
\end{tabular}%

\caption{Results for semantic experiments in a \textbf{few-shot} setting. }
\label{tab:task-b-few-shot}
\end{table*}

\begin{table*}[ht]
\centering
\begin{tabular}{llll}
\textbf{Model} &
  \textbf{\begin{tabular}[c]{@{}l@{}}F-1 Score\\ EN\end{tabular}} &
  \textbf{\begin{tabular}[c]{@{}l@{}}F-1 Score\\ PT\end{tabular}} &
  \textbf{\begin{tabular}[c]{@{}l@{}}F-1 Score\\ ALL\end{tabular}} \\ \hline
GPT-4           & $\textbf{0.6552}\pm 0.031$ & $0.5955\pm 0.031$          & $\textbf{0.6392}\pm 0.025$ \\
GPT-3           & $0.6231\pm0.061$           & $\textbf{0.6056}\pm 0.041$ & $0.6225\pm 0.050$          \\
Llama 3-70B     & $0.5514\pm 0.037$          & $0.5600\pm 0.018$          & $0.4289\pm 0.030$          \\
Llama 3-8B      & $0.5696\pm 0.076$          & $0.5677\pm 0.077$          & $0.5689\pm 0.026$          \\ \hline
Random baseline & 0.5441                     & 0.5170                     & 0.5387                    \\ \hline
\end{tabular}%

\caption{Experiment results for \textit{detection experiments} in the \textbf{zero-shot} setting using \textbf{synthetic MWEs.}}
\label{tab:task-a-zero-shot-synthetic}
\end{table*}

\begin{table*}[ht]
\centering
\begin{tabular}{llll}
\textbf{Model} &
  \textbf{\begin{tabular}[c]{@{}l@{}}F-1 Score\\ EN\end{tabular}} &
  \textbf{\begin{tabular}[c]{@{}l@{}}F-1 Score\\ PT\end{tabular}} &
  \textbf{\begin{tabular}[c]{@{}l@{}}F-1 Score\\ ALL\end{tabular}} \\ \hline
GPT-4           & $0.5454\pm 0.016$         & $0.4855\pm 0.025$          & $0.5183\pm 0.017$          \\
GPT-3           & $\textbf{0.5655}\pm0.004$ & $\textbf{0.5204}\pm 0.071$ & $\textbf{0.5444}\pm 0.036$ \\
Llama 3-70B     & $0.4655\pm 0.028$         & $0.3885\pm 0.030$          & $0.4289\pm 0.030$          \\
Llama 3-8B      & $0.5455\pm 0.029$         & $0.4711\pm 0.090$          & $0.5108\pm 0.060$          \\ \hline
Random baseline & 0.5092                    & 0.4985                     & 0.5025                    \\ \hline
\end{tabular}%
\caption{Experiment results for \textit{semantic experiments} in the \textbf{zero-shot} setting using \textbf{synthetic MWEs.}}
\label{tab:synthetic-task-b}
\end{table*}

\section{Results and Discussion}
\label{results}
\paragraph{Are LLMs effective in interpreting idiomatic MWEs?}
In general, LLM abilities in detecting idiomatic expressions fall short of smaller fine-tuned PLMs.
This is especially true for Galician and Portuguese experiments using the original dataset. 
Given that most of these models are predominantly trained on English data, it is unsurprising that they perform best on the English portion of the dataset, with most models reaching an F1 score of 0.70+ in the zero-shot and few-shot settings. 
Interestingly, the overall macro F1 score shows a slight improvement for all models in the CS experiments. 
This improvement is likely due to the inclusion of Spanish in all examples, a language that is likely more prevalent in the pre-training data compared to Galician and Portuguese. 
Some models exceed xlm-roBERTa's F1 score for the English partition of the data, 0.7070, but none are able to reach a score higher than 0.76. 
However, all models fall short of the baseline for Portuguese, 0.6803, and the overall F1 score, 0.6540, when compared to a fine-tuned xlm-RoBERTa-base model in both zero-shot and few-shot settings. 
Notably, the Llama3-8B-instruct model failed to produce valid outputs in the few-shot setting for both the original and CS datasets.
It seems that the combination of few-shot examples along with the input text confounded the model, to the degree that most of the model outputs were not a single word answer containing the label. 
Figures~\ref{fig:f1-original-dataset} and~\ref{fig:f1-cs-dataset} demonstrate how Llama3-8B struggled in few-shot experiments, evidenced by the high variance in the scores in the three prompts used to generate model output. 
Although LLMs have been competitive in detecting nuanced language when compared to a much smaller PLM, in general, they struggle in the other lower-resourced languages. 

\paragraph{To what extent do LLMs capture the non-compositional meanings of MWEs?}
LLMs seem to struggle to capture the meaning of text containing MWEs. 
In general, unlike in the MWE interpretation task, the few-shot examples here helped most models reach higher F1 scores for both Portuguese and English text.
GPT-4 was able to perform comparably to the xlm-roBERTa baseline for English in the few-shot setting, with a score of 0.7638, marginally beating xlm-roBERTa's score of 0.7590. 
All models fall short of the combined baseline score and F1 score in Portuguese in zero-shot and few-shot settings.
This may indicate that models are not capturing the true meaning of text containing MWEs and may be relying on spurious correlations to make decisions. 
In general, it seems that LLMs struggle to capture the meaning of nuanced language, given that they struggle to surpass a fine-tuned PLM with many fewer parameters.

\paragraph{What is the impact of providing information in the prompt?}
The results indicate LLMs generally perform above random on synthetic MWE interpretation task, were we provided the definition of the MWE in the prompt, with GPT-4 achieving the highest combined F1 score of 0.6392.
However, their performance is notably lower compared to when real MWEs are used. 
This suggests that models may have memorised some real MWEs during training. 
For the \textit{semantic experiments} involving synthetic MWEs, model performance remains close to the random baseline.
Across all languages, GPT-3 achieves the highest scores, 0.5444, but these results still do not significantly exceed the random baseline of 0.5025.
This demonstrates that providing additional information in the prompt does not improve model performance when the vocabulary differs from patterns encountered during pre-training.
The models appear to rely heavily on statistical patterns from pre-training data, and when faced with unfamiliar patterns, they fail to generate correct responses, even with contextual cues. 
Overall, the models have difficulty applying the meanings of synthetic MWEs to complete the task, suggesting that additional context does not help them address complex linguistic issues or unfamiliar vocabulary.
 
\paragraph{Is there a relationship between LLM size and performance on text containing non-compositional phrases?}
The smaller Llama 8B model performed unexpectedly well on the zero-shot detection task, achieving results comparable to Llama 70B and GPT-3 for English text. 
However, it showed a slight decline in performance for Portuguese and Galician. 
Notably, the Llama 8B model was unable to process both the original text as well as the CS text in the few-shot detection task and was sensitive more sensitive to prompting than the other models, evidenced by the high standard deviation seen in Table~\ref{tab:idiom-detection-few-shot}.
GPT-4 outperformed other LLMs in English for the semantic task and was the only LLM capable of handling English-Spanish CS text in this context. 
While further experiments are necessary, these findings suggest that a smaller model, like Llama 8B, may suffice to achieve results comparable to much larger models for MWE detection in English.
However, the results also show that GPT-4 is better than other models when confronted with unknown vocabulary and representing the semantics of MWEs compared to the other LLMs.

\subsection{Error Analysis}
Certain MWEs were consistently classified correctly by all models when using the original dataset for the MWE interpretation task, as detailed in Table~\ref{tab:correct_MWEs}. 
Figures~\ref{fig:f1-original-dataset} and~\ref{fig:f1-cs-dataset} illustrate the sensitivity of models to different prompts. 
Notably, the standard deviations, especially for Llama 8B, show how different some model outputs may be depending on the prompt.
A subset of MWEs was correctly classified by all models in the detection task, these are listed in Table~\ref{tab:correct_MWEs}.
It lists correctly identified MWEs by all models, those correctly classified by specific model families, and the overlap between the two largest models. 
We also examined correlations between the error rates of individual MWEs across models. 
There is a moderate correlation of 0.64 between the error rates of Llama models, and a strong correlation of 0.70 between Llama 70B and GPT-4.
However, the correlation between the two GPT models is weaker, at 0.32, possibly indicating differences in their training data.
Figure~\ref{fig:errors_mwe} shows the errors between Llama 70B and GPT-4 as well as the two Llama models. 
The curves are overlaid for comparison.  
The strong correlation in errors between certain models may indicate shared limitations in tasks requiring nuanced language understanding, potentially stemming from their architectural design or training data.

\section{Conclusion}
\label{conclusion}
Our study examines whether state-of-the-art LLMs can effectively capture and encode the meanings of nuanced language, specifically through MWEs, across various languages and cultural contexts.
Through a series of experiments in English, Portuguese, Galician, and code-switched text, we find that LLMs are not fully capable of detecting and representing the semantics of MWEs. 
While their performance in English is comparable to that of a smaller fine-tuned PLM for the detection task, they underperform in other languages and tasks. 
Interestingly, CS text supports the models in metalinguistic tasks, but hinders their ability to represent the meaning of MWEs.
This is highlighted in the semantic experiments, where it is easier for the model to detect if MWEs are used as idioms than to determine and capture the meanings of text containing idioms.
Our error analysis highlights the limitations of current models with handling text containing challenging linguistic phenomena like MWEs. 
Looking forward, we suggest exploring open models that match the performance of GPT-4, focusing on methods to improve the models’ ability to capture the meanings of nuanced language.
New approaches are required to enable models to effectively address challenging linguistic phenomena, including MWE and CS.

\section{Limitations}
Our experiments are conducted on models primarily trained on English-language data.
Although many multilingual models include some training data in other languages, most of their training remains in English.
However, our results show that the models do not perform as expected, even on English text.
In some experiments, we use synthetic CS data instead of naturally occurring CS data.

\bibliographystyle{unsrt}  
\bibliography{references}  

\appendix
\label{sec:appendix}

\section{Prompts and Examples}
\label{app:prompts}

\begin{table*}[ht]
\resizebox{\textwidth}{!}{%
\begin{tabular}{ll}
\textbf{Prompts}    &        \textbf{Type}               \\
\hline
You are an expert linguist in code-switching between \{lang\} and Spanish (CS-lang-name). \\
You will translate sentences you get into code-switch by switching some of the words into Spanish. \\
Make sure that there is a balance between two languages.\\
You will translate everything into code-switch apart from the multiword expression, denoted as 'MWE'. & Contains MWE        \\
\hline
You are an expert linguist in code-switching between \{lang\} and Spanish (CS-lang-name). \\
You will translate sentences you get into code-switch by switching some of the words into Spanish. \\
Make sure that there is a balance between two languages. You will translate everything into code-switch.                                                      & Adversarial example \\
\hline
\end{tabular}%
}
\caption{These are the prompts used to generate the synthetic CS dataset. The term lang corresponds to either English, Portuguese, or Galician. CS-lang-name is the code-switched language name between Spanglish, or Portuñol. This part of the prompt is nor included when asking for Galician Spanish mix. }
\label{tab:CS-prompts}
\end{table*}

\begin{table*}[ht]
\resizebox{\textwidth}{!}{%
\begin{tabular}{ll|l}
\textbf{Prompts} &
  \textbf{Type} &
  Task \\  \hline
\begin{tabular}[c]{@{}l@{}}You are a linguist specializing in phraseology with extensive knowledge of idiomatic expressions, colloquialisms, and figurative language.\\ Your task is to identify whether a given text contains a multiword expression used figuratively as an idiom.\\ First, understand the meaning of the text provided by the user. Then, determine if the multiword expression is used figuratively or literally.\\ Respond with 'idiom' for figurative use or 'literal' for literal use.\\ Provide a one-word answer.\end{tabular} &
Think &
Detection \\\cline{1-2}
\begin{tabular}[c]{@{}l@{}}You are a linguist specializing in phraseology.\\With your deep knowledge of idiomatic expressions, colloquialisms, and figurative language, you are considered the best in the field. \\Instruction: Given a text containing a multiword expression, determine if it is used as an idiom or literal.\\ Provide a one word answer.\end{tabular} &
Persona \\ \cline{1-2}
\begin{tabular}[c]{@{}l@{}}Analyze the provided text to determine whether the given Multiword Expression (MWE) is used as an idiom or literally.\\ If the MWE is used figuratively as an idiom, answer "Idiom"; if it is used in its literal sense, answer "Literal".\\ Provide a one word answer.\end{tabular}  &
  chatGPT &
   \\ \hline
\begin{tabular}[c]{@{}l@{}}You are a linguist specializing in phraseology with extensive knowledge of idiomatic expressions, colloquialisms, and figurative language.\\ Your task is to identify whether two sentences have a similar meaning or not.\\ Some of the sentences may contain idiomatic expressions.\\ First, understand the meaning of the text provided by the user.\\ Then, determine if the sentences have a similar meaning.\\ Respond with 'true' if the sentences are similar in meaning or 'false' if they mean different things .\\ Provide a one-word answer\end{tabular} &
  Think &
  Semantic \\\cline{1-2}
\begin{tabular}[c]{@{}l@{}}You are a linguist specialising in semantics. You possess a deep knowledge of language structure, and meaning.\\ You are the best in the field.\\ You will be presented with two sentences, some of the sentences contain figurative language.\\ You will determine if two sentences are similar in meaning.\\ Output 'false' if their meanings are different or 'true' if their meanings are similar.\\ Provide a one word answer\end{tabular} &
  Persona &
   \\ \cline{1-2}
\begin{tabular}[c]{@{}l@{}}Analyze the provided text to determine if the given sentences, which may include idiomatic expressions, have similar meanings.\\ If the sentences are similar, output "true." If the meanings differ, output "false."\\ Provide a one-word answer. answer\end{tabular} &
  chatGPT &
   \\ \hline
\end{tabular}%
}
\caption{These are the system prompts provided to all models for each experiment, categorised by whether the experiment falls under 'Detection' or 'Semantic'.}
\label{tab:prompts-experiments}
\end{table*}
\begin{table*}[ht]
\resizebox{\textwidth}{!}{%
\begin{tabular}{lll}
\textbf{Examples of Data Inputs to Models} &
  \textbf{MWE} &
  \textbf{Experiment} \\ \hline
But we need to remember, it was a very close call. &
  Close call &
  Detection \\\hline
\begin{tabular}[c]{@{}l@{}}Takes about tres años for the last milk tooth to come in después de que el first one sprouts. (es-en-mix)\\ Takes about three years for the last milk tooth to come in after the first one sprouts. (original)\end{tabular} &
  Milk tooth &
  CS \\\hline
\begin{tabular}[c]{@{}l@{}}Sentence 1 : Witten's swan song was far from a hit.\\ Sentence 2: Witten's final performance was far from a hit.\end{tabular} &
  Swan song &
  Semantic \\\hline
Despite not receiving an inheritance, Eve Jobs is still living the wrawpths chaumb. &
  \textit{\begin{tabular}[c]{@{}l@{}}Wrawpths chaumb\\ replaces High life\end{tabular}} &
  Synthetic MWE
\end{tabular}%
}
\caption{Examples of different text inputs provided to the model for different experiments. }
\label{tab:examples_data}
\end{table*}

\begin{table*}[h]
\resizebox{\textwidth}{!}{%
\begin{tabular}{llll}
\textbf{Correct for all} & \textbf{Correct for Llama} & \textbf{Correct for GPT} & \textbf{Correct for GPT-4 and Llama 70B} \\
arma branca   & arma branca         & arma branca   & arma branca \textbullet~inner product \\
camisa branca & auditoría externa   & camisa branca & auditoría externa \textbullet~lime tree      \\
centros educativos       & camisa branca              & centros educativos       & incendios forestais  \textbullet~lingua galega       \\
exame clínico & centros educativos  & dirty word    & baby buggy       \textbullet~market place \\
glóbulos vermellos       & enerxía limpa              & glóbulos vermellos       & benign tumour \textbullet~net income          \\
              & exame clínico       & horas baixas  & black cherry    \textbullet~news agency \\
              & glóbulos vermellos  & news agency   & camisa branca    \textbullet~nut case      \\
              & incendios forestais & olho gordo    & centros educativos \textbullet~olho gordo      \\
              & lingua galega       & pavio curto   & exame clínico                       \\
              & voto secreto        &               & glóbulos vermellos                  \\

\end{tabular}%
}
\caption{The first column demonstrates the overlap between the MWEs the models got correct all the time. The second column shows the overlap of correct words between the two Llama models, column three shows the overlap between the GPT models, and the last column shows the overlap between GPT-4 and Llama 70B.}
\label{tab:correct_MWEs}
\end{table*}

 
\end{document}